\documentclass[11pt,a4paper]{article}
\usepackage[hyperref]{acl2021}
\usepackage{times}
\usepackage{latexsym}

\usepackage{microtype}
\usepackage{adjustbox}
\usepackage{latexsym}
\usepackage{multirow}
\usepackage{diagbox}
\usepackage{amsmath}
\usepackage{pifont}
\usepackage{graphicx}
\usepackage{amssymb}
\usepackage{arydshln}

\usepackage{booktabs}
\usepackage{caption}
\usepackage{subcaption}
\usepackage{color}
\usepackage{makecell}
\usepackage{xcolor}
\usepackage{algorithm}
\usepackage[noend]{algpseudocode}
\usepackage{pifont}
\usepackage{amsmath}

\usepackage{kotex}

\definecolor{cusorange}{RGB}{255, 50, 50}
\definecolor{medblue}{RGB}{28, 98, 215}
\definecolor{darkgreen}{RGB}{0, 100, 0}
\definecolor{darkred}{RGB}{167, 51, 51}

\newcommand{\cls}{[\textsc{cls}]}
\newcommand{\sep}{[\textsc{sep}]}
\newcommand{\mask}{[\textsc{mask}]}
\newcommand{\rel}{[\textsc{rel}]}

\aclfinalcopy
\def\aclpaperid{2231}

\title{Deep Context- and Relation-Aware Learning \\for Aspect-based Sentiment Analysis}

\makeatletter
\newcommand{\printfnsymbol}[1]{%
  \textsuperscript{\@fnsymbol{#1}}%
}
\makeatother

\author{Shinhyeok Oh$^{1}\thanks{\ \ These two authors equally contributed to this work.}\;\, \thanks{\ \ This work was done while the author was an intern at Kakao Enterprise.}$\,, Dongyub Lee$^{2}\printfnsymbol{1}$, Taesun Whang$^{3}$, Ilnam Park$^{4}$\textbf{,} \\~\textbf{Gaeun Seo}$^{4}$\textbf{,} \textbf{Eunggyun Kim}$^{4}$\textbf{, and Harksoo Kim}$^{5}\thanks{\ \ Corresponding author.}$\\
$^{1}$Netmarble AI Center \quad $^{2}$Kakao Corp. \quad $^{3}$Wisenut Inc.\\ $^{4}$Kakao Enterprise \quad $^{5}$Konkuk University\\
}

\date{}

\begin{document}
\maketitle
\begin{abstract}
Existing works for aspect-based sentiment analysis (ABSA) have adopted a unified approach, which allows the interactive relations among subtasks. However, we observe that these methods tend to predict polarities based on the literal meaning of aspect and opinion terms and mainly consider relations implicitly among subtasks at the word level. In addition, identifying multiple aspect--opinion pairs with their polarities is much more challenging. Therefore, a comprehensive understanding of contextual information w.r.t. the aspect and opinion are further required in ABSA. In this paper, we propose Deep Contextualized Relation-Aware Network (DCRAN), which allows interactive relations among subtasks with deep contextual information based on two modules (i.e., Aspect and Opinion Propagation and Explicit Self-Supervised Strategies). Especially, we design novel self-supervised strategies for ABSA, which have strengths in dealing with multiple aspects. Experimental results show that DCRAN significantly outperforms previous state-of-the-art methods by large margins on three widely used benchmarks.
\end{abstract}
\section{Introduction}

Aspect-based sentiment analysis (ABSA) is a task of identifying the sentiment polarity of associated aspect terms in a sentence. Generally, ABSA is composed of three subtasks, 1) aspect term extraction (ATE), 2) opinion term extraction (OTE), and 3) aspect-based sentiment classification (ASC). Given the sentence ``\textit{Food is good, but service is dreadful.}", ATE aims to identify two-aspect terms ``\textit{food}" and ``\textit{service}", and OTE aims to determine two-opinion terms ``\textit{good}" and ``\textit{dreadful}". Then, ASC assigns a sentiment polarity of each aspect: ``\textit{food} (\textit{positive})" and ``\textit{service} (\textit{negative})".

Existing works for ABSA have adopted a two-step approach, which considers each subtask separately~\cite{tang2016aspect, xu2018double}. However, most recently, unified approaches have achieved significant performance improvements in ABSA task. \citet{luo2020grace} focused on modeling the interactions between aspect terms and \citet{chen2020relation} exploited dyadic and triadic relations between subtasks (i.e., ATE, OTE, ASC).

\begin{table}[t] \centering
\renewcommand{\arraystretch}{1.1}
\begin{adjustbox}{width=0.48\textwidth}
\resizebox{1.0\columnwidth}{!}{
\begin{tabular}{clccc}
\toprule

& Examples (Ground Truth)  & Model & Aspect (Polarity) & Opinion \\
\midrule
\multirow{2}{*}{E1} & \multirow{2}{*}{\begin{tabular}[c]{@{}l@{}}I’ve had \textit{\textcolor{medblue}{better}} \textit{\textcolor{darkgreen}{Japanese food}} ({\textcolor{darkred}{\text{neg}}})\\ at a mall food court.\end{tabular}}
      & RACL  & Japanese food (pos)  & better \\
      \cdashline{3-5}
&     & DCRAN & Japanese food (neg)  & better \\
        
\midrule
\multirow{2}{*}{E2} & \multirow{2}{*}{\begin{tabular}[c]{@{}l@{}}The \textit{\textcolor{darkgreen}{sushi}} ({\textcolor{darkred}{\text{neg}}}) is cut in blocks\\
\textit{\textcolor{medblue}{bigger}} than my cell phone. \end{tabular}}
& RACL  & sushi (neu)  & bigger \\
\cdashline{3-5}
&     & DCRAN & sushi (neg)  & bigger\\
\midrule

\multirow{5}{*}{E3}
& \multirow{5}{*}{\begin{tabular}[c]{@{}l@{}}
While the \textit{\textcolor{darkgreen}{smoothies}} ({\textcolor{darkred}{\text{neg}}}) are a little 
\\\textit{\textcolor{medblue}{bigger}} for me, the \textit{\textcolor{darkgreen}{fresh juices}} ({\textcolor{darkred}{\text{pos}}}) 
\\are the \textit{\textcolor{medblue}{best}} I have ever had !\end{tabular}}
& \multirow{3}{*}{RACL}
& \multirow{3}{*}{\begin{tabular}[c]{@{}c@{}}smoothies (pos)\\ fresh juices (pos)\end{tabular}}
& \multirow{3}{*}{\begin{tabular}[c]{@{}c@{}}bigger\\ fresh\\ best\end{tabular}}\\
&    &     &   &  \\
&    &     &   &  \\
\cdashline{3-5}
&  & \multirow{2}{*}{DCRAN}    &  \multirow{2}{*}{\begin{tabular}[c]{@{}c@{}}smoothies (neg)\\ fresh juices (pos)\end{tabular}}  
& \multirow{2}{*}{\begin{tabular}[c]{@{}c@{}}bigger\\ best\end{tabular}}\\
&  &  &  & \\

\bottomrule

\end{tabular}
}
\end{adjustbox}
\caption{Examples of ABSA results comparing to previous approach~\cite{chen2020relation} that we reimplement. All the results are based on BERT$_{base}$ model for a fair comparison. The polarity labels pos, neu, and neg, denote positive, neutral, and negative, respectively. 
}
\label{table:introduction}
\vspace{-0.2cm}

\end{table}

Despite the impressive results, their methods have two limitations. First, they only consider relations among subtasks at the word level and do not explicitly utilize contextualized information of the whole sequence. For example, E1 in Table~\ref{table:introduction}, the opinion term ``\textit{better}" seems to represent positive opinion of ``\textit{Japanese food}". However, the authentic meaning of E1 is ``\textit{The Japanese food I have had at the food court was more delicious than the one I had at this restaurant}". Thus, previous approaches tend to assign polarities based on the literal meaning of aspect and opinion terms (E2). Second, identifying multiple aspect--opinion pairs and their polarities is much more challenging as the model needs to not only detect multiple aspects and opinions but also correctly predict each polarity of the aspect (E3).

To address the aforementioned issues, we propose Deep Contextualized Relation-Aware Network (DCRAN) for ABSA. DCRAN not only implicitly allows interactive relations among the subtasks of ABSA, but also explicitly considers their relations by using contextual information. Our main contributions are as follows: 1) We design aspect and opinion propagation decoder so that the model has a comprehensive understanding of the whole context, and thus it results in better prediction of the polarity. 2) We propose novel self-supervised strategies for ABSA, which are highly effective in dealing with multiple aspects and considering deep contextualized information with the aspect and opinion terms. To the best of our knowledge, it is the first attempt to design explicit self-supervised methods for ABSA. 3) Experimental results demonstrate that DCRAN significantly outperforms previous state-of-the-art methods on three widely used benchmarks.

\section{DCRAN: Deep Contextualized Relation-Aware Network}

\begin{figure*}[t]\centering
\includegraphics[width=0.85\textwidth]{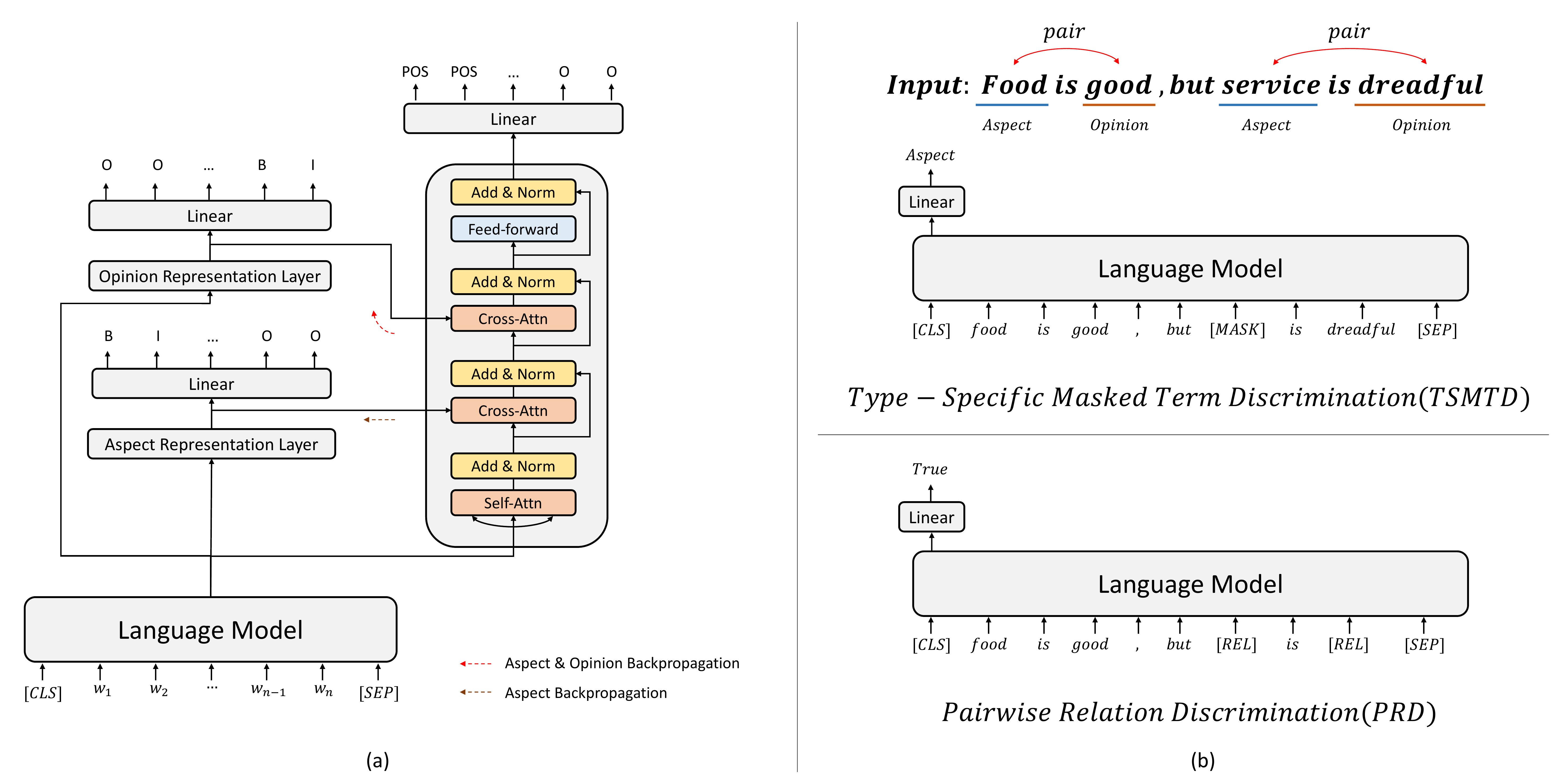}
\caption{Overall architecture of Deep Contextualized Relation-Aware Network (DCRAN) for ABSA.}
\label{fig:appendix_model_architecture}
\vspace{-0.2cm}
\end{figure*}

\subsection{Task Definition}
Given a sentence $S = \{w_1, w_2, ..., w_n\}$, where \( n\) denotes the number of tokens, we aim to solve three subtasks: aspect term extraction (ATE), opinion term extraction (OTE), and aspect-based sentiment classification (ASC) as sequence labeling problems. ATE task aims to identify a sequence of aspect terms \( Y^{a} = \{y^{a}_{1}, y^{a}_{2}, ...,  y^{a}_{n}\}\), where \( y^{a}_{i} \in \{B, I, O\} \), and OTE task aims to identify a sequence of opinion terms \( Y^{o} = \{y^{o}_{1}, y^{o}_{2}, ...,  y^{o}_{n}\}\), where \( y^{o}_{i} \in \{B, I, O\} \) of aspect and opinion terms, respectively. Likewise, ASC task aims to assign a sequence of polarities \( Y^{p} = \{y^{p}_{1}, y^{p}_{2}, ...,  y^{p}_{n}\}\), where \( y^{p}_{i} \in \{POS, NEU, NEG, O\} \). The labels \textit{POS}, \textit{NEU}, and \textit{NEG} denote \textit{positive}, \textit{neutral}, and \textit{negative}, respectively.

\subsection{Task-Shared Representation Learning} 
\label{section:task-shared}
Following existing works, we utilize pre-trained language models, such as BERT~\cite{devlin2018bert} and ELECTRA~\cite{clark2020electra} as the shared encoder to construct context representation, which is shared by subtasks: ATE, OTE, and ASC. Given a sentence $S = \{w_1, w_2, ..., w_n\}$, pre-trained language models take the input sequence, \(\mathbf{X}_{\text{absa}} = [\cls\,w_1\,w_2\,...\,w_n\,\sep]\), and output a sequence of the shared context representation, \(H = \{h_{\cls}, h_1, h_2, ..., h_n, h_{\sep}\} \in \mathbb{R}^{d_h\times (n+2)}\), where \(d_h\) represents a dimension of the shared encoder. We represent the parameters of the shared encoder as \(\Theta_{s}\). Then, we utilize a single-layer feed-forward neural network (FFNN) as,
\begin{equation}
\label{eq:1}
\begin{split}
Z^{a} &= (W_{1}h_{[1:n+1]} + b_{1})\\
\hat{Y}^{a} &= \text{softmax}(W_{2}Z^{a} + b_{2}), 
\end{split}
\end{equation}
where \(W_{1} \in \mathbb{R}^{d_h\times d_h}\) and \(W_{2} \in \mathbb{R}^{3\times d_h}\) are trainable parameters. The parameters of a single-layer FFNN are represented as \(\Theta_{a}\) for aspect term extraction. The objective of aspect term extraction is minimizing the negative log-likelihood (NLL) loss: \(\mathcal{L}_{\text{ate}}(\Theta_{s}, \Theta_{a}) = - \sum \log p(Y^{a}|H).\)
Likewise, \(Z^{o}\) and  \(\hat{Y}^{o}\) are obtained as in Equation~\ref{eq:1}. Then, the NLL loss of opinion term extraction is defined as, \(\mathcal{L}_{\text{ote}}(\Theta_{s}, \Theta_{o}) = - \sum \log p(Y^{o}|H).\)

\subsection{Aspect and Opinion Propagation} 
We utilize the transformer-decoder~\cite{vaswani2017attention} to consider relations of aspect and opinion while predicting a sequence of polarities. Our transformer-decoder is mainly composed of a multi-head self-attention, two multi-head cross attention, and a feed-forward layer. The multi-head self-attention takes shared context representation \(H\) as,
\begin{equation}
U^{h} = \text{LN}(H + \text{SelfAttn}(H, H, H))
\end{equation}
and \(U^{h}\), \( Z^{a}\), and  \(Z^{o}\)  are fed into two steps of cross multi-head attention as,
\begin{align}
U^{a} &= \text{LN}(U^{h} + \text{CrossAttn}(U^{h}, Z^{a}, Z^{a})) \label{eq3}\\
U^{o} &= \text{LN}(U^{a} + \text{CrossAttn}(U^{a}, Z^{o}, Z^{o})) \label{eq4}
\end{align}
where LN represents layer norm~\cite{ba2016layer}. Note that Equation~\ref{eq3} and~\ref{eq4} represent aspect and opinion propagation, respectively. Then \(U^{o}\) is fed into a single-layer FFNN to obtain a sequence of polarities \(Y^{p}\). The objective of aspect-based sentiment analysis is minimizing the NLL loss: \( \mathcal{L}_{\text{asc}}(\Theta_{s}, \Theta_{a}, \Theta_{o}, \Theta_{p}) = - \sum \log p(Y^{p}|H, Z^{a}, Z^{o}).\) The architecture of the aspect and opinion propagation is described in Figure~\ref{fig:appendix_model_architecture}-(a).
 
\subsection{Explicit Self-Supervised Strategies} 
\label{sec3:explicit}
To further exploit the aspect--opinion relation with contextualized information of a sentence, we propose explicit self-supervised strategies consisting of two auxiliary tasks: 1) type-specific masked term discrimination (TSMTD) and 2) pairwise relations discrimination (PRD). The examples of \textit{Explicit Self-Supervised Strategies} are described in Figure~\ref{fig:appendix_model_architecture}-(b).

\paragraph{Type-Specific Masked Term Discrimination} In the type-specific masked term discrimination task, we uniformly mask aspects, opinions, and terms that do not correspond to both, using the special token \(\mask\). The input sequence of a masked sentence is represented as, $\mathbf{X}_{\text{tsmtd}} = [\cls\,w_1\,...\,\mask_i\,...\,w_n\,\sep]$, and is fed into pre-trained language models. Then, the output representation of \(\cls\) token is used to classify which type of term is masked in a sentence as,

\begin{alignat*}{2}
\hat{Y}^{m} &= \text{softmax}(W_{3}h_{\cls} + b_{3}),
\end{alignat*}
where \(W_{3} \in \mathbb{R}^{3 \times d_h}\) represents trainable parameters and \(\hat{Y}^{m} \in \{Aspect, Opinion, O\}\). The parameters of a linear projection layer are represented as \(\Theta_{m}\) for the type-specific masked term discrimination. Then, the NLL loss of the type-specific masked term discrimination is defined as: \( \mathcal{L}_{\text{tsmtd}}(\Theta_{s}, \Theta_{m}) = - \sum \log p(Y^{m}|H). \)

This allows the model to explicitly exploit sentence information by discriminating what kind of term is masked.

\paragraph{Pairwise Relations Discrimination} In this task, we uniformly replace both aspects and opinion terms using the special token \(\rel\). The input sequence of a masked sentence is represented as, $\mathbf{X}_{\text{prd}} =[\cls\,w_1\,...\,\rel_i\,...\,\rel_j\,...\,w_n\,\sep]$, and is fed into pre-trained language models. Then, the output representation of \(\cls\) token is used to discriminate whether the replaced tokens have a pairwise relation as,
\begin{alignat*}{2}
\hat{Y}^{r} &= \text{softmax}(W_{4}h_{\cls} + b_{4}),
\end{alignat*}
where \(W_{4} \in \mathbb{R}^{2 \times d_h}\) represents trainable parameters and \(\hat{Y}^{r} \in \{True, False\}\). The parameters of a linear projection layer are represented as \(\Theta_{r}\) for the pairwise relations discrimination. Then, the NLL loss of the pairwise relations discrimination is defined as: \(\mathcal{L}_{\text{prd}}(\Theta_{s}, \Theta_{r}) = - \sum \log p(Y^{r}|H).\)

We describe the negative sampling method to replace aspects and opinion terms in Appendix~\ref{appendix:sampling}.

\subsection{Joint Learning Procedure}
All these tasks are jointly trained, and the final objective is defined as,
\begin{alignat*}{2}
\mathcal{L}_{\text{absa}} &= \mathcal{L}_{\text{ate}} + \mathcal{L}_{\text{ote}} + \mathcal{L}_{\text{asc}} \\
\mathcal{L}_{\text{aux}} &= \mathcal{L}_{\text{tsmtd}} + \mathcal{L}_{\text{prd}} \\
\mathcal{L}_{\text{final}} &= \mathcal{L}_{\text{absa}} + \alpha\mathcal{L}_{\text{aux}}
\end{alignat*}
where \(\alpha\) is a hyper-parameter determining the degree of auxiliary tasks. Note that the parameters \(\Theta_{s}\) are optimized for all subtasks. Especially, the parameters \(\Theta_{s}\) are further optimized through \(\mathcal{L}_{\text{tsmtd}}\) and \(\mathcal{L}_{\text{prd}}\) to explicitly exploit the relations between aspect and opinion with context meaning.

\section{Experiments}
\begin{table*}[t]

\centering
\resizebox{1.0\textwidth}{!}{
\begin{tabular}{|l|l|cccc|cccc|cccc|}
\hline
\multicolumn{2}{|l|}{\multirow{2}{*}{}}    & \multicolumn{4}{c|}{LAP14}   & \multicolumn{4}{c|}{REST14}         & \multicolumn{4}{c|}{REST15} \\ 
\multicolumn{2}{|l|}{} & ATE-F1 & OTE-F1 & ASC-F1 & ABSA-F1 & ATE-F1 & OTE-F1 & ASC-F1 & ABSA-F1 & ATE-F1 & OTE-F1 & ASC-F1 & ABSA-F1 \\ \hline
MNN~\cite{wang2018towards}                                        & GloVe & 76.94 & 77.77 & 65.98 & 53.80 & 83.05 & 84.55 & 68.45 & 63.87 & 70.24 & 69.38 & 57.90 & 56.57\\
E2E-TBSA~\cite{li2019unified}               & GloVe         & 77.34  & 76.62  & 68.24  & 55.88    & 83.92  & 84.97  & 68.38  & 66.60   & 69.40  & 71.43  & 58.81  & 57.38   \\
DOER~\cite{luo2019doer}                                       & GloVe & 80.21 & - & 60.18 & 56.71 & 84.63 & - & 64.50 & 68.55 & 67.47 & - &36.76 & 50.31 \\
IMN$^{-d}$~\cite{he2019interactive}               & GloVe         & 78.46 & 78.14 & 69.62 & 57.66 &84.01 &85.64 &71.90 &68.32 &69.80 &72.11 & 60.65 & 57.91 \\
RACL~\cite{chen2020relation}                & GloVe       & 81.99 & 79.76 &71.09 & 60.63 & 85.37 & 85.32 & 74.46 & 70.67 &72.82 & 78.06 & 68.69 & 60.31\\
WHW~\cite{peng2020knowing}                   & GloVe          & -      & 74.84  & -      & 62.34  & -      & 82.45  & -      & 71.95   & -      & 78.02  & -      & 65.79    \\
\hline
IKTN~\cite{liang2020iterative}               & BERT$_{base}$   & 80.89   & 78.90  & 73.42  & 62.34 & 86.13  & \underline{86.62}  & 74.35  & 71.75   & 71.63  & \underline{76.79}  & 69.85  & 62.33      \\
SPAN~\cite{hu2019open}                       & BERT$_{large}$  & 82.34   & -      & 62.50  & 61.25 & \underline{86.71}  & -      & 71.75  & 73.68   & \underline{74.63}  & -      & 50.28  & 62.29      \\
IMN$^{-d}$~\cite{he2019interactive}                & BERT$_{large}$  & 77.55   & \textbf{81.00}  & 75.56  & 61.73 & 84.06  &85.10   & 75.67  & 70.72   & 69.90  & 73.29  & 70.10  & 60.22 \\
Dual-MRC~\cite{mao2021joint}                & BERT$_{large}$  & \underline{82.51}   & -      & \underline{75.97}  & \underline{65.94} & 86.60  & -      & \textbf{82.04}  & \underline{75.95}   & \textbf{75.08}  & -   & 73.59  & 65.08     \\
RACL~\cite{chen2020relation}                & BERT$_{large}$  & 81.79 & \underline{79.72}  & 73.91    & 63.40 & 86.38  & \textbf{87.18}  & \underline{81.61}  & 75.42   & 73.99  & 76.00  & \underline{74.91}  & \underline{66.05}    \\
\hline
{\multirow{4}{*}{DCRAN (Ours)}}             & BERT$_{base}$ & 81.76  & 78.84  & 77.02  & 65.18  & 88.21  & 86.36  & 78.67  & 75.77   & 71.61  & 75.86  & 73.30  & 63.19    \\
                                            & BERT$_{large}$ & \textbf{83.40}	& \underline{79.72}	& \textbf{78.75}	& \textbf{68.07} & \textbf{88.73} & 86.07 & 80.64 & \textbf{77.28} & 74.45 & \textbf{78.45}	& \textbf{76.30}	& \textbf{67.92} \\
                                            \cdashline{2-14}
                                            & ELECTRA$_{base}$ & \textbf{85.69}  & \textbf{80.19}  & 79.36  & 70.22  & 89.64  & 87.30  & 84.12  & 80.00   & 77.41 & 78.80  & \textbf{78.55}  & 71.67   \\
                                            & ELECTRA$_{large}$ &85.61 &79.77 &\textbf{80.78} &\textbf{71.47} & \textbf{89.67} &\textbf{87.59} & \textbf{84.22} &\textbf{80.32} &\textbf{79.68} &\textbf{79.90} &77.99 &\textbf{73.67} \\
\hline
\end{tabular}}
\caption{Evaluation results on the LAP14, REST14, and REST15 datasets, which are provided by~\citet{chen2020relation}. All the results except ours are cited from the existing works \cite{chen2020relation,peng2020knowing,mao2021joint} and all the baselines are described in Appendix \ref{appendix:baselines}. We report average results over five runs with random initialization. The best scores are in bold, and the second-best scores are underlined depending on the types of the pre-trained language model. `-' denotes unreported results. 
} 
\label{table:main_results}
\vspace{-0.2cm}
\end{table*}

\subsection{Experimental Setup}
We evaluate our model on three widely used sentiment analysis benchmarks: laptop reviews (LAP14), restaurant reviews (REST14) from~\cite{pontiki2014semeval}, and restaurant reviews (REST15) from~\cite{pontiki2015semeval}. Primitive versions of these benchmarks only provide aspect terms and sentiment polarities, while opinion terms are provided by~\citet{wang2016recursive, wang2017coupled} later. Recently,~\citet{fan2019target} provides aspect-opinion pairwise datasets (Section~\ref{sec3:explicit}). Following \citet{he2019interactive}, we set four evaluation metrics: ATE-F1, OTE-F1, ASC-F1, and ABSA-F1. The ATE-F1, OTE-F1, and ASC-F1 measure each subtask's F-1 scores, and ABSA-F1 measures complete ABSA, which counts only when both ATE and ASC predictions are correct.  

\subsection{Quantitative Results}
Table~\ref{table:main_results} reports the quantitative results on the LAP14, REST14, and REST15 datasets. Our experiments utilize two pre-trained language models such as BERT and ELECTRA, for the shared encoder. First, we observe that DCRAN-BERT$_{base}$ shows slightly lower ABSA-F1 scores than previous state-of-the-art methods, which is based on BERT$_{large}$, on the REST14 and LAP14 datasets except for the REST15 dataset. This suggests that our proposed methods are highly effective for ABSA. Overall, DCRAN-BERT$_{large}$ significantly outperforms previous state-of-the-art methods in all metrics. Another observation is that ELECTRA based models outperform BERT based models. As a result, DCRAN-ELECTRA$_{large}$ achieves absolute gains over previous state-of-the-art results by 5.5\%, 4.4\%, and 7.6\% in ABSA-F1 on the LAP14, REST14, and REST15 datasets, respectively.

\begin{table}[t]\centering
\begin{adjustbox}{width=0.45\textwidth}

\begin{tabular}{llc}
\toprule
\multicolumn{2}{l}{} & ABSA-F1 \\
\midrule
 & DCRAN-ELECTRA$_{base}$   & \textbf{80.00}$^{\dagger}$  \\
\midrule
\multirow{3}{*}{\begin{tabular}[c]{@{}l@{}}
Aspect and Opinion \\ Propagation
\end{tabular}}
& \quad\,  w/o AP           & 79.44$^{\dagger}$   \\
& \quad\,  w/o OP           & 79.58$^{\dagger}$   \\
& \quad\,  w/o AP \& OP     & 79.08$^{\dagger}$   \\
\midrule
\multirow{3}{*}{\begin{tabular}[c]{@{}l@{}}
Explicit Self-Supervised \\ Strategies
\end{tabular}}
& \quad\,  w/o TSMTD          & 79.56$^{\dagger}$  \\
& \quad\,  w/o PRD            & 79.40$^{\dagger}$  \\
& \quad\,  w/o TSMTD \& PRD   & 79.03$^{\dagger}$  \\
\midrule

\multirow{2}{*}{\begin{tabular}[c]{@{}l@{}}
Baseline
\end{tabular}} 
& \multirow{2}{*}{\begin{tabular}[c]{@{}l@{}}
\quad\,  w/o \& AP \& OP \\ \qquad\;\;\,\, \& TSMTD \& PRD \end{tabular}}
& \multirow{2}{*}{\begin{tabular}[c]{@{}l@{}}
78.61 \end{tabular}}  \\
& & \\
\bottomrule
\end{tabular}

\end{adjustbox}

\caption{Ablation study on the REST14 dataset. We choose DCRAN-ELECTRA$_{base}$ as the baseline. $\dagger$ denotes statistical significance (p-value $<$ 0.05).}
\label{table:ablation}
\vspace{-0.2cm}
\end{table}
\begin{table*}[t]\centering
\small
\resizebox{0.9\textwidth}{!}{\begin{tabular}{ll|cc|cc}
\hline
& & \multicolumn{2}{c|}{REST14} & \multicolumn{2}{c}{REST15} \\
& & ABSA-F1  & Sent-level Acc.   & ABSA-F1  & Sent-level Acc.     \\ \hline
\multirow{3}{*}{\begin{tabular}[c]{@{}l@{}}
Single-\\Aspect
\end{tabular}} & DCRAN\_ELECTRA$_{base}$              & \textbf{78.62} & \textbf{74.48} & \textbf{66.23} & \textbf{67.69} \\
                   & \ \ \ w/o TSMTD \& PRD                   & 78.42  & 73.79 & 64.21 & 66.67 \\
                   & \ \ \ w/o TSMTD \& PRD \& AP \& OP       & 77.45  & 73.10 & 62.50 & 64.29 \\ \hline
\multirow{3}{*}{\begin{tabular}[c]{@{}l@{}}
Multiple-\\Aspect
\end{tabular}} & DCRAN\_ELECTRA$_{base}$           & \textbf{81.19} & \textbf{64.24} & \textbf{68.20} & \textbf{52.34} \\
                   & \ \ \ w/o TSMTD \& PRD                 & 80.22 & 61.70 & 65.16 & 48.60 \\
                   & \ \ \ w/o TSMTD \& PRD \& AP \& OP     & 79.88 & 61.39 & 64.84 & 46.73 \\
\bottomrule

\end{tabular}}
\caption{Aspect analysis on the REST14 and REST15 datasets. Comparisons of ABSA-F1 and sentence-level accuracy results for the case when the sentence contains single-aspect or multiple-aspect.}
\label{table:analysis}
\vspace{-0.2cm}
\end{table*}

\subsection{Ablation Study}
To study the effectiveness of the aspect propagation (AP), opinion propagation (OP), type-specific masked term discrimination (TSMTD), and pairwise relations discrimination (PRD), we conduct ablation experiments on the REST14 dataset. We set the baseline model that did not utilize aspect and opinion propagation and explicit self-supervised strategies. When the AP and OP are not utilized, a single-layer FFNN is utilized as in Equation~\ref{eq:1} to predict a sequence of polarities \(Y^{p}\) instead of transformer-decoder. As shown in Table~\ref{table:ablation}, we can observe that the AP is more effective than the OP, and scores drop significantly when not utilizing the AP and OP. In the case of explicit self-supervised strategies, we can observe that the PRD is more effective than the TSMTD. As the PRD objective is discriminating whether the replace tokens have a pairwise aspect--opinion relations, it allows the model to more exploit the relations between aspect and opinion at a sentence level. 

\subsection{Aspect Analysis}

We conduct aspect analysis by comparing sentences with single- and multiple-aspect. As shown in Table~\ref{table:analysis}, \textit{Aspect and Opinion Propagation} significantly improves performance when the sentence contains a single-aspect, while a small increase is observed w.r.t. the case of multiple-aspect. Although considering the relations between aspect and opinion implicitly can improve performance w.r.t. the case of single-aspect, it is not sufficient for inducing performance improvement for the multiple-aspect case. It suggests that additional explicit tasks are further required to identify multiple-aspect with corresponding opinions, which helps the model assign polarities correctly. In the case of multiple-aspect, \textit{Explicit Self-Supervised Strategies} show absolute ABSA-F1 improvements of 0.97\% (80.22\% $\rightarrow 81.19\%$) and 3.04\% (65.16\% $\rightarrow 68.20$) on the REST14 and REST15 datasets, respectively. This indicates explicit self-supervised strategies are highly effective for correctly identifying ABSA when the sentence contains multiple-aspect. In addition, the performance gain by \textit{Explicit Self-Supervised Strategies} in Table~\ref{table:ablation} is mostly derived from the multiple-aspect cases (+0.97\%), thus our proposed model has strengths in dealing with multiple aspects.

In ABSA, it is important to accurately predict all aspects and corresponding sentiment polarities in a sentence. Since ABSA-F1 is a word-level based metric, it still has a limitation to evaluate whether all aspects and corresponding polarities are correct or not. Therefore, we also evaluate our method with sentence-level accuracy; the number of sentences that accurately predicted all aspects and polarity in a sentence divided by total number of sentences. Unlike ABSA-F1, the sentence-level accuracy of multiple-aspect is lower than that of single-aspect, which implies identifying multiple aspects and their polarities is more challenging. In the case of multiple-aspect, our \textit{Explicit Self-Supervised Strategies} leads significant sentence-level accuracy improvements of 2.54\% (61.70\% $\rightarrow 64.24\%$) and 3.74\% (48.60\% $\rightarrow 52.34\%$) on the REST14 and REST15 datasets, respectively. However, we observe only small improvements in sentence-level accuracy on both datasets when the sentence contains single-aspect. From these observations, we demonstrate that our proposed method is highly effective for the case when the sentence contains multiple aspects.

\section{Conclusion}
In this paper, we proposed the Deep Contextualized Relation-Aware Network (DCRAN) for aspect-based sentiment analysis. DCRAN allows interaction between subtasks implicitly in a more effective manner and two explicit self-supervised strategies for deep context- and relation-aware learning. We obtained the new state-of-the-art results on three widely used benchmarks.

\section*{Acknowledgements}
We thank the anonymous reviewers, Dongsuk Oh, Jungwoo Lim, and Heuiseok Lim for their constructive comments. This work was partially supported by Institute of Information \& communications Technology Planning \& Evaluation(IITP) grant funded by the Korea government(MSIT)(No.2020-0-00368, A Neural-Symbolic Model for Knowledge Acquisition and Inference Techniques) and supported by the National Research Foundation of Korea(NRF) grant funded by the Korea government(MSIT) (No. 2020R1F1A1069737).

\bibliographystyle{acl_natbib}
\bibliography{acl2021}

\clearpage
\appendix
\renewcommand{\thesection}{\Alph{section}}
\renewcommand{\thesubsection}{\Alph{section}.\arabic{subsection}}
\section{Appendix}
\subsection{Related Work}
Existing works have studied a two-step approach for ABSA. In a two-step approach, each model for ATE, OTE, and ASC are separately trained and are merged in a pipelined manner~\cite{wang2016recursive, tang2016aspect, wang2017coupled, he2017unsupervised, xu2018double, yu2018global, li2018aspect, chen2019transfer}. However, the errors from other tasks can be propagated to the ASC and can degrade performance after all. 

Most recently, a unified approach that comprised of joint approach~\cite{garcia2018w2vlda, luo2019doer, he2019interactive, luo2020grace} and collapsed approach~\cite{li2017learning, ma2018joint, wang2018towards, li2019unified} has been proposed. A joint approach labels each word with different tag sets for each task: ATE, OTE, and ASC. On the other hand, a collapsed approach labels each word as the combined one of ATE and ASC, such as ``B-POSITIVE" and ``I-POSITIVE", where ``B" and ``I" represent the aspect term boundary, and ``POSITIVE" represents polarity. However, in a collapsed approach, the relations among subtasks cannot be effectively exploited because subtasks need to share all representation without distinction of each task. Therefore, a joint training approach allows the interactive relations between subtasks, while a collapsed approach is not.

\subsection{Negative Sampling Algorithm for Pairwise Relations Discrimination}
Algorithm \ref{alg:algorithm1} describes the negative sampling procedure in pairwise relations discrimination. The get\_sample function takes a list of aspect-opinion pairs in a sentence and replaces them with \(\rel\) tokens. Then, if the replaced tokens have pairwise relations, set the target label as True, and set as False if not. The get\_pair function randomly selects a pairwise aspect and opinion, and the get\_neg\_pair function selects aspects and opinions of different pairs when there are two or more pairs in a sentence.

\label{appendix:sampling}
\algrenewcommand\algorithmicrequire{\textbf{Input:}}
\algrenewcommand\algorithmicensure{\textbf{Output:}}

\begin{algorithm}[t]
    \caption{Negative Sampling Algorithm for Pairwise Relations Discrimination}
    \begin{algorithmic}
    \Require $pairs$: list of aspect--opinion pairs in a sentence
    \Ensure $pair$, $target$
    \Function {Get\_Sample}{$pairs$}
        \If{count($pairs$) == 0}
        \State return None, None
        \ElsIf{count($pairs$) == 1}
        \State return $pairs$[0], True
        \Else{}
        \State $random$ = $\{0 {<} random \leq 1 \}$
        \EndIf
        \If{$random \leq 0.25$}
        \State return get\_pair($pairs$), True
        \Else{}
        \State return get\_neg\_pair($pairs$), False
        \EndIf
    \EndFunction
    \end{algorithmic}
    \label{alg:algorithm1}
\end{algorithm}

\subsection{Implementation Details}
We implemented our model by using the PyTorch \cite{adam2019pytorch} deep learning library based on the open source\footnote{\begin{footnotesize}https://github.com/huggingface/transformers\end{footnotesize}} (i.e., Transformers \cite{wolf2020transformers}). For the shared encoder, we adopt four types of pre-trained language models: BERT$_{base}$, BERT$_{large}$, ELECTRA$_{base}$, and ELECTRA$_{large}$. We set the batch size to 64 for the $base$ model, 12 for the BERT$_{large}$ and 32 for the ELECTRA$_{large}$. We set the initial learning rate to 5e-5 for BERT$_{base}$ and ELECTRA$_{base}$, 2e-5 for BERT$_{large}$, and 5e-6 for ELECTRA$_{large}$. For the transformer decoder, we set the number of heads in multi-head attention and hidden layers to 2 among range from 2 to 6, and hidden dimension size to 768. In the case of \(\alpha\), we obtained the best results when \(\alpha\) is 1. The average runtime for each approach was about 20 seconds for BERT$_{base}$ and ELECTRA$_{base}$, and 90 seconds for BERT$_{large}$ and ELECTRA$_{large}$. We train our models using AdamP \cite{heo2021adamp} optimizer and conduct experiments with Tesla V100 GPU for all the experiments. 

\begin{table*}[!hbtp] 
\centering
\resizebox{0.8\textwidth}{!}{
\begin{tabular}{clccc}
\toprule

& Examples (Ground Truth)  & Model & Aspect (Polarity) & Opinion \\
\midrule
\multirow{3}{*}{E1} & \multirow{3}{*}{\begin{tabular}[c]{@{}l@{}}I have worked in restaurants and cook a lot,\\and there is no way a maggot should be able\\to get into \textit{\textcolor{medblue}{well prepared}} \textit{\textcolor{darkgreen}{food}} ({\textcolor{darkred}{\text{neg}}}).\end{tabular}}
      & RACL  & food (pos)  & well \\
      \cdashline{3-5}
&     & DCRAN w/o & food (pos)  & well prepared \\
      \cdashline{3-5}
&     & DCRAN & food (neg) & well prepared \\
     
\midrule
\multirow{5}{*}{E2} & \multirow{5}{*}{\begin{tabular}[c]{@{}l@{}}All in all, I would return - as it was a \textit{\textcolor{medblue}{beautiful}}\\ \textit{\textcolor{darkgreen}{restaurant}} ({\textcolor{darkred}{\text{pos}}}) - but I hope the \textit{\textcolor{darkgreen}{staff}} ({\textcolor{darkred}{\text{neg}}})\\pays more attention to the little details in the future.\end{tabular}}
      & RACL  & -  & - \\
      \cdashline{3-5}
&  & \multirow{2}{*}{DCRAN w/o}    &  \multirow{2}{*}{\begin{tabular}[c]{@{}c@{}}restaurant (pos)\\ staff (pos)\end{tabular}}  
& \multirow{2}{*}{beautiful}\\ \\
      \cdashline{3-5}
&  & \multirow{2}{*}{DCRAN}    &  \multirow{2}{*}{\begin{tabular}[c]{@{}c@{}}restaurant (pos)\\ staff (neg)\end{tabular}}  
& \multirow{2}{*}{beautiful}\\ \\

\midrule
\multirow{6}{*}{E3}
& \multirow{6}{*}{\begin{tabular}[c]{@{}l@{}}
I have never been so \textit{\textcolor{medblue}{disgusted}} by both \\ \textit{\textcolor{darkgreen}{food}} ({\textcolor{darkred}{\text{neg}}}) and \textit{\textcolor{darkgreen}{service}} ({\textcolor{darkred}{\text{neg}}}) \end{tabular}}
& \multirow{2}{*}{RACL}    &  \multirow{2}{*}{\begin{tabular}[c]{@{}c@{}} food (pos)\\ service (pos)\end{tabular}}
& \multirow{2}{*}{disgusted} \\ \\
      \cdashline{3-5}
&  & \multirow{2}{*}{DCRAN w/o}    &  \multirow{2}{*}{\begin{tabular}[c]{@{}c@{}}food (pos)\\ service (neg)\end{tabular}}  
& \multirow{2}{*}{disgusted}\\ \\
      \cdashline{3-5}
&  & \multirow{2}{*}{DCRAN}    &  \multirow{2}{*}{\begin{tabular}[c]{@{}c@{}}food (neg)\\ service (neg)\end{tabular}}  
& \multirow{2}{*}{disgusted}\\ \\
\bottomrule

\end{tabular}
}
\caption{{Case study on the REST15 dataset. Model comparison between  previous state-of-the-art method (RACL)~\cite{chen2020relation} and our proposed method (DCRAN). DCRAN w/o denotes DCRAN without \textit{Explicit Self-Supervised Strategies} (Section~\ref{sec3:explicit}). All models are built based on the {BERT$_{base}$}} model. The polarity labels pos, neu, and neg denote positive, neutral, and negative, respectively. `-' denotes that the model failed to extract corresponding terms.}
\label{table:intro_compare}
\vspace{-0.3cm}

\end{table*}
\subsection{Baselines}
\label{appendix:baselines}
We compare our model with the following previous works\footnote{\begin{footnotesize}We do not compare our work with GRACE~\cite{luo2020grace} as \citet{luo2020grace} contains \textit{conflict} tag in polarities.\end{footnotesize}}.
\paragraph{MNN~\cite{wang2018towards}} is a multi-task model for ATE and ASC using attention mechanisms to learn the joint representation of aspect and polarity relations.
\vspace{-0.1cm}
\paragraph{E2E-TBSA~\cite{li2019unified}} is an end-to-end model of the collapsed approach for ATE and ASC. Additionally, it introduces the auxiliary OTE task without explicit interaction.
\vspace{-0.1cm}
\paragraph{DOER~\cite{luo2019doer}} is a dual cross-shared RNN framework that jointly trains ATE and ASC. It considers relations between aspect and polarity.
\vspace{-0.1cm}
\paragraph{IMN~\cite{he2019interactive}} is a multi-task model for ATE and ASC with separate labels. The OTE task is fused into ATE by constructing five-class labels. 
\vspace{-0.1cm} 
\paragraph{WHW~\cite{peng2020knowing}} is a unified two-stage framework to extract (aspect, opinion, polarity) triples as a result of ATE, OTE, and ASC.
\vspace{-0.1cm}
\paragraph{IKTN~\cite{liang2020iterative}} is an iterative knowledge transfer network for ABSA considering the semantic correlations among the ATE, OTE, and ASC. 
\vspace{-0.1cm}
\paragraph{SPAN~\cite{hu2019open}} is a pipeline approach to solve ATE and ASC using {BERT$_{large}$}. It uses a multi-target extractor for ATE and a polarity classifier for ASC. 
\vspace{-0.1cm}
\paragraph{RACL~\cite{chen2020relation}} defines interactive relations among ATE, OTE, and ASC. It proposes relation propagation mechanisms through the stacked multi-layer network. 
\vspace{-0.1cm}
\paragraph{Dual-MRC~\cite{mao2021joint}} leverages two machine reading comprehension problems to solve ATE and ASC. It jointly trains two BERT-MRC models sharing parameters. 

\subsection{Case Study}
\label{appendix:case_study}
In E1 and E3, while all models correctly extract both aspect and opinion, RACL and DCRAN w/o make inaccurate polarities predictions based on the words having superficial meaning (i.e., \textit{well prepared}, \textit{disgusted}). Especially, E3 expresses a sarcastic opinion about aspect terms throughout the sentence. It suggests that these models cannot understand the authentic meaning of the sentence. On the other hand, DCRAN grasps the entire context and predicts the correct polarity corresponding to its aspect. In E2, the evidence for understanding the actual meaning of the aspect term \textit{staff} is not specified in a word-level opinion and expressed in a sentence like ``\textit{I hope the staff pays more attention to the little details in the future}''. In this case, RACL can not extract aspect and opinion terms, and DCRAN w/o make inaccurate polarities predictions for the aspect term \textit{staff} based on the opinion term \textit{beautiful}. However, DCRAN with \textit{Explicit Self-Supervised Strategies} understands the sentence expressing an opinion on the \textit{staff} and predicts correctly.

\end{document}